\documentclass[11pt]{article}
\usepackage{acl}

\usepackage{times}
\usepackage{latexsym}

\usepackage[T1]{fontenc}

\usepackage[utf8]{inputenc}

\usepackage{microtype}

\usepackage{inconsolata}

\usepackage{graphicx}

\usepackage{booktabs}   
\usepackage{multirow}   
\usepackage{graphicx}   

\usepackage{algorithm}
\usepackage{algorithmic}
\usepackage{amsmath} 

\title{Deconstructing Spatial Complexity: Hierarchical Decomposition\\for LLM Spatial Reasoning}

\author{Yi Wang, Haojie Lu, Zhaofan Zhang, Li Chen, \and Sihong Xie \\
  The Hong Kong University of Science and Technology (Guangzhou) \\
  \texttt{ywang778@connect.hkust-gz.edu.cn}}

\begin{document}
\maketitle
\begin{abstract}
LLMs have shown remarkable proficiency in general language understanding and reasoning. However, they consistently underperform in spatial reasoning that severely limits their application, particularly in embodied intelligence. Inspired by the success of hierarchical reinforcement learning, this paper introduces a novel method for hierarchical task decomposition in LLM spatial reasoning. Our approach guides LLMs to decompose complex tasks into manageable sub-tasks by identifying key intermediate states and generating simplified sub-environments. However, we identify that LLMs often fail to derive optimal intermediate states due to their insufficient spatial prior, leading to sub-optimal task decomposition. To address this limitation and enhance its planning capability, we propose the MCTS-Guided Group Relative Policy Optimization (M-GRPO), where we reformulate the UCT formula by incorporating the LLM’s prior predictive probabilities alongside its epistemic uncertainty. Furthermore, we implement a more fine-grained advantage function, enabling the model to learn optimal path planning. Experimental results demonstrate that our method substantially improves LLM performance on spatial tasks, including navigation, planning, and strategic games, achieving state-of-the-art results. This work paves the way for LLMs in real-world applications.
\end{abstract}

\begin{figure}[t] 
    \centering
    \includegraphics[width=1.0\columnwidth]{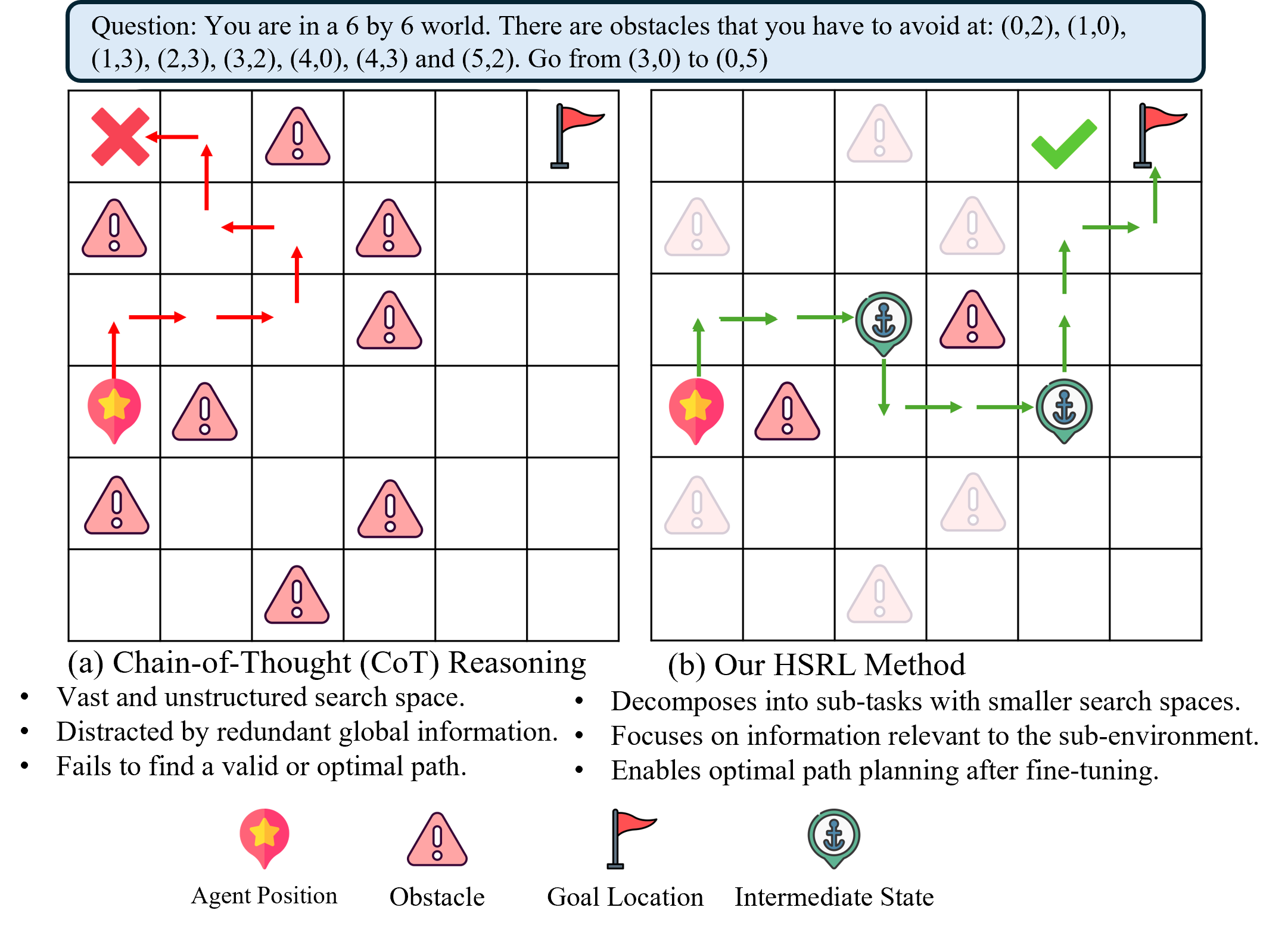} 
    \caption{Comparison between CoT Reasoning and our HSRL method. \textbf{(a)} Standard CoT reasoning fails on complex spatial planning by inefficiently exploring a vast search space amidst distracting information. \textbf{(b)} In contrast, our HSRL framework succeeds by decomposing the task via key intermediate states and constructing focused sub-environments, which enables efficient and optimal planning.}
    \label{fig:figure1}
\end{figure}

\section{Introduction}

Large Language Models (LLMs) have revolutionized the landscape of artificial intelligence, achieving remarkable breakthroughs across various domains, including natural language processing and scientific reasoning \citep{zhao2023survey}. However, as LLMs transition into the era of embodied AI, a critical and persistent bottleneck has emerged: their inherent limitations in spatial reasoning. While LLMs excel at manipulating abstract concepts and language, they often struggle with understanding complex spatial relationships, performing efficient path planning, and engaging in sequential action reasoning\citep{ma2025explorllm,chen2024autotamp}. This severely limits their development and practical deployment in embodied systems.

Existing research has explored several avenues to address this challenge, yet each faces significant limitations. Prompt engineering methods like CoT\citep{wei2022chain}, ToT\citep{yao2023tree} and ProgPrompt\citep{singh2023progprompt} aim to elicit reasoning through specialized prompts, but their effectiveness is capped by the model's often flawed intrinsic spatial capabilities. Fine-tuning approaches \citep{dao2025alphamaze,deng2025can,aghzal2024look} show promise but typically demand vast, expensive task-specific datasets and suffer from poor generalization to novel environments. Task decomposition strategies like HyperTree \citep{gui2025hypertree} and Plan-and-Act \citep{erdogan2025plan} are primarily designed for tasks with clear, language-based "logical breaks". For example, "cooking" naturally includes sub-steps like "buying ingredients" and "chopping." In such tasks, sub-goals have distinct semantic boundaries, allowing models to decompose them using pre-trained semantic priors. Consequently, these strategies are rendered ill-suited for spatial reasoning problems like pathfinding that lack such linguistic segmentation. Finally, offloading planning to external, non-differentiable tools breaks the end-to-end optimization paradigm, as these tools cannot be jointly trained with the LLM's representation layer and may not be universally deployable at test time.

To overcome these limitations, we introduce Hierarchical Spatial Reasoning with LLM (HSRL). The core innovation of HSRL lies in its state- and environment-based hierarchical mechanism, which is fundamentally different from prior language-based decomposition methods. The framework employs a two-level hierarchy. A high-level LLM planner decomposes complex tasks by generating a structured sequence of intermediate states, which serve as sub-goals to guide the overall planning process. Then, for each state-to-state transition, we decouple the process into two distinct components: first, an Environment Processor constructs a localized sub-environment, and subsequently, a Low-Level Action Generator generates the precise actions required to reach the target sub-goal. While this hierarchical structure provides a powerful framework for decomposition, the high-level planner's effectiveness is constrained by the LLM's insufficient spatial priors, leading to the generation of sub-optimal intermediate states. To address this, we propose an innovative online fine-tuning framework, M-GRPO, designed to enhance the high-level planner. Our approach improves planning optimality by tackling two fundamental challenges: effective exploration of the solution space and precise credit assignment for training. To achieve robust exploration, we draw inspiration from Monte Carlo Tree Search (MCTS), where the high-level LLM generates multiple candidate sequences of intermediate states, building a search tree to systematically explore diverse planning strategies. Unlike standard LLM-MCTS integration paradigms \citep{xie2monte, guanrstar} rely on the vanilla UCT formula, we propose incorporating the LLM's intrinsic confidence and uncertainty as internal reward signals. This integration facilitates the selection of trajectories that exhibit both high environmental rewards and high internal certainty, thereby mitigating the instability caused by high-reward but low-confidence paths—which often represent stochastic hallucinations that hinder convergence.\citep{kuhn2023semantic,azaria2023internal} Furthermore, we leverage perplexity-based uncertainty as an informative metric to evaluate whether the model has sufficiently explored the state space within its semantic knowledge, leading to a more principled and efficient search process. For precise credit assignment, we introduce a fine-grained advantage function, a significant departure from traditional Group Relative Policy Optimization (GRPO) which evaluates whole-trajectory values without detailed supervision. Our method calculates the advantage of each intermediate state relative to its ”sibling“ states (i.e., those that share a common prefix state sequence). This provides a focused and accurate training signal, enabling the LLM to learn which specific sub-goals are the most effective. Our method requires only a small amount of data and can be flexibly applied to multi-level planning tasks. In summary, this work makes the following key contributions:

\begin{itemize}
    \item \textbf{A Novel State- and Environment-Based Hierarchical Reasoning Framework:} We introduce HSRL, a framework that presents a novel state- and environment-based decomposition paradigm for LLM spatial reasoning, departing from prevalent language-based methods. This paradigm is specifically designed to address continuous spatial problems where traditional language-based decomposition is ineffective.
    \item \textbf{A Novel Fine-Tuning Framework for Planning Optimality:} To address the sub-optimal planning inherent in pre-trained LLMs, we develop M-GRPO, a new fine-tuning algorithm. By integrating a modified Monte Carlo Tree Search exploration mechanism with a fine-grained, node-level advantage function, our method substantially improves planning optimality with high data efficiency. 
    \item \textbf{Comprehensive Empirical Validation of Superiority:} Through extensive experiments on large-scale navigation, object planning, and strategy game benchmarks, we demonstrate that HSRL achieves state-of-the-art performance. The results validate its significant gains over existing methods and its strong generalization across diverse task modalities. 
\end{itemize}

\begin{figure*}[t] 
    \centering
    \includegraphics[width=1\textwidth]{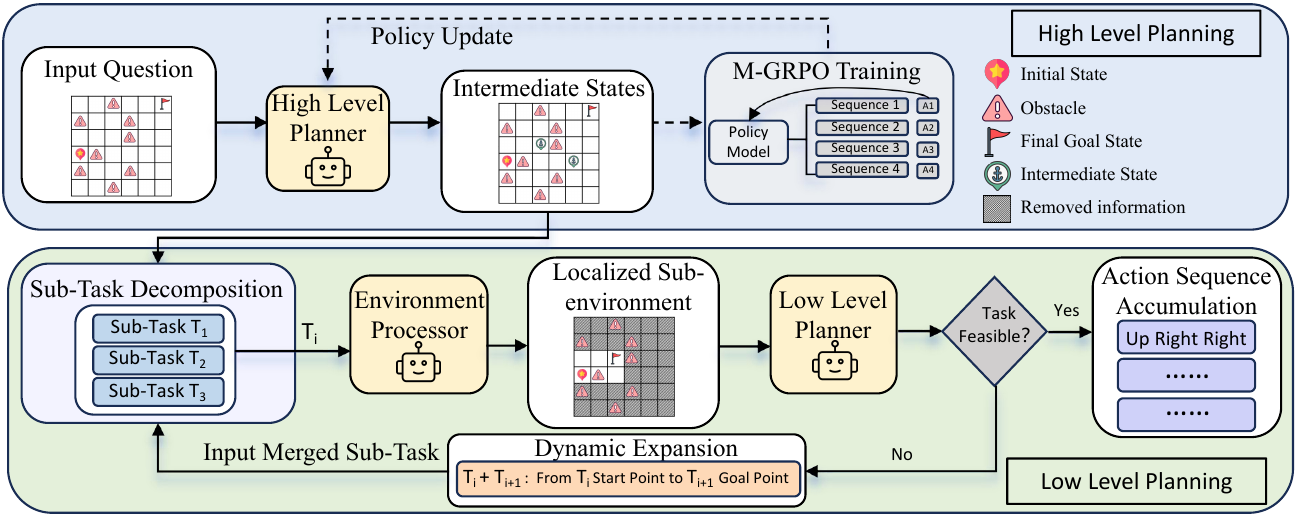} 
     \caption{An overview of the HSRL framework. This framework employs a two-level hierarchical strategy. An M-GRPO trained high-level planner first identifies key intermediate states, decomposing the task into a series of sub-tasks. A low-level planner selects relevant information for the sub-task and then generates action sequences for each sub-task within a localized sub-environment. If a sub-task is unsolvable, it is merged with the next one (e.g., from the start of Subtask 1 to the end of Subtask 2) and replanned.}
    \label{fig:method}
\end{figure*}

\section{Method}

In this work, we introduce the HSRL framework, as illustrated in Figure \ref{fig:method}, to address the limitations of existing LLM-based planning methods (see Appendix \ref{gen_inst}). Our approach consists of two key components: a novel two-level hierarchical framework that decomposes complex tasks into a series of manageable sub-problems, and an innovative MCTS-guided finetuning method designed to enhance the optimality of the generated plans.

\subsection{Hierarchical Planning with State and Environment Decomposition}

Our framework leverages a two-level hierarchical decomposition strategy to break down complex planning tasks. This decomposition is applied at both the state level and the environmental level, effectively managing the complexity of the problem space.

\paragraph{State-Level Decomposition via LLM.} Prior research in LLM-based path planning has shown promising results by manually decomposing tasks into sub-goals \citep{aghzal2024look}. We extend this concept by enabling the LLM to autonomously generate these key intermediate states. Given a task's initial and final states, our method prompts the LLM to reason and generate a concise sequence of critical intermediate states. This process transforms a high-level goal into a series of state-to-state transitions, effectively simplifying the planning horizon for subsequent steps.

\paragraph{Environmental-Level Decomposition and Dynamic Expansion.} After decomposing the task at the state level, much of the global environmental information becomes irrelevant noise for solving a specific sub-task, which can hinder the reasoning process. Following the generation of the state sequence, we define a sub-task for each consecutive pair of intermediate states. For each sub-task, we create a corresponding regional environment by identifying information that is relevant to the current sub-problem (e.g., obstacles or landmarks within a localized area). This hierarchical representation allows the model to focus on a smaller, more manageable sub-environment, thereby improving efficiency and reducing the search space. If the model is unable to find a valid path within the localized environment, the scope of the sub-task is expanded. The end state is extended to the next intermediate state in the sequence, creating a larger sub-task that encompasses a broader area. This process is repeated until a solution is found or, in the worst case, the problem reverts to the original, full-scale task, ensuring robust and complete coverage of the problem space.

\begin{figure*}[t] 
    \centering
    \includegraphics[width=0.9\textwidth]{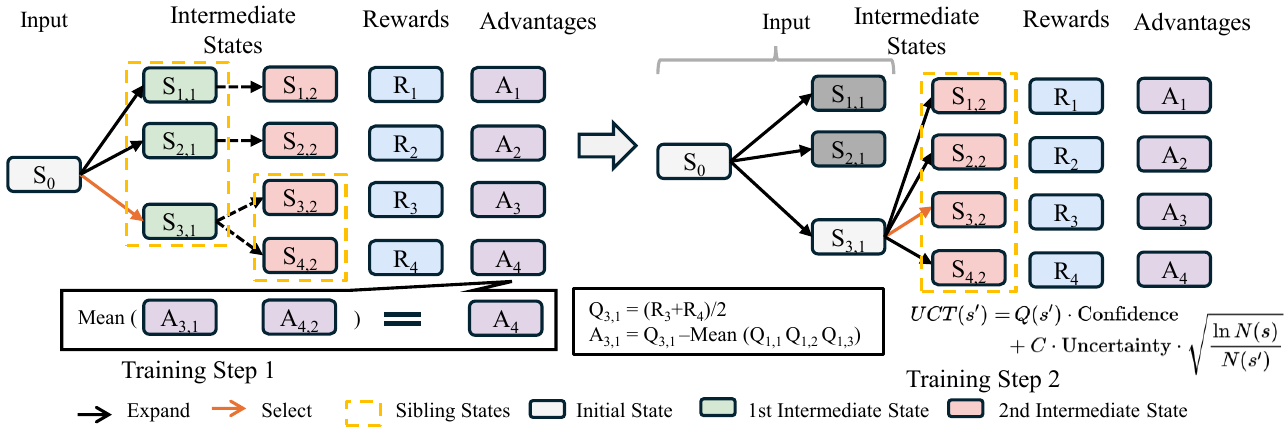} 
    \caption{Overview of the M-GRPO algorithm. Starting from the initial state $S_0$, the framework expands four simulation sequences in parallel. The \textbf{advantage values} of nodes within these sequences are calculated to guide the model training step. Post-training, the \textbf{pivotal intermediate waypoint} is selected based on the reformulated UCT formula, serving as the root for the subsequent expansion and training cycle. This iterative process continues until the MCTS termination criterion is satisfied.}
    \label{fig:m-grpo}
\end{figure*}

\subsection{Optimizing Planning with M-GRPO}

Due to an inherent lack of pre-training in spatial reasoning, LLMs often struggle to generate optimal sequences of intermediate states. However, the generation of correct intermediate states is a critical prerequisite for the success of subsequent environment decomposition and low-level action planning. To address this, we propose an online learning approach, as illustrated in Figure \ref{fig:m-grpo}, that integrates the exploratory power of MCTS with the fine-tuning process of GRPO. This approach enables the LLM to learn and improve its planning policy during exploration.

\paragraph{MCTS-Guided Exploration with LLM-Induced Prior Preference.}
We formulate intermediate-state generation as a tree search problem, where each node denotes a partial reasoning state and each root-to-leaf path defines a candidate trajectory. Starting from the initial state, the search tree is iteratively expanded through selection, expansion, simulation, and backpropagation.

A central challenge is how to incorporate the LLM's internal confidence into the search process without treating it as a calibrated probability of correctness. To this end, we use the generation likelihood of the newly fine-tuned LLM policy to construct a prior preference over candidate states. For a generated child state $s'$ with token sequence $x_{1:|T|}$, we compute the average log-likelihood:
\begin{equation}
\ell(s')=\frac{1}{|T|}\sum_{t=1}^{|T|}\log P(x_t\mid x_{<t},\text{context}).
\end{equation}
Based on this quantity, we define the LLM-induced prior confidence as
\begin{equation}
c(s')=\exp(\tau \cdot \ell(s')),
\end{equation}
where $\tau$ denotes the temperature parameter. Intuitively, candidates that are more semantically coherent and more supported by the model receive a larger weight in the exploitation term. Let $Q(s')$ denote the empirical value estimate of state $s'$, computed from Monte Carlo returns, then the prior-modulated value estimate can be obtained by $\tilde{Q}(s') = c(s')Q(s')$, where the empirical node value is calibrated by an LLM-induced prior.

Meanwhile, to encourage exploration toward low-likelihood regions where the LLM policy is less certain, we define an exploration factor
\begin{equation}
u(s') = 1 + \gamma \cdot \mathrm{clamp}(-\ell(s'), 0, u_{\max}),
\end{equation}
which serves as an uncertainty-aware exploration amplifier. Larger uncertainty leads to a stronger exploration bonus, encouraging the search to gather additional evidence in the model's epistemically weak regions.

We then define the refined selection score as
\begin{equation} \label{eq:final_ucb}
U(s';s)=\tilde{Q}(s') + C\,u(s')\sqrt{\frac{\ln N(s)}{N(s')}},
\end{equation}
considering exploitation and exploration for key points' distribution and further LLM rollout.
The next state is selected by
\begin{equation}
s_{\text{next}}=\arg\max_{s'\in\mathrm{Children}(s)} U(s';s).
\label{eq:final_ucb}
\end{equation}
By considering the uncertainty and LLM prior preference over plausible candidates, this formulation augments the standard optimism-under-uncertainty structure of UCTs.

\begin{algorithm}[t]
\small
\caption{M-GRPO Training Algorithm}
\label{alg:m-grpo-functional}
\begin{algorithmic}[1]
\REQUIRE Policy $\pi_{\theta}$, initial state $s_0$, iterations $N_{\text{max}}$, group size $M$
\ENSURE Optimized policy $\pi_{\theta}$

\STATE $T \leftarrow \text{InitializeTree}(s_0)$
\FOR{it = 1 \TO $N_{\text{max}}$}
    \STATE \textit{// MCTS-Guided Exploration}
    \STATE $L \leftarrow \text{SELECTLEAF}(T)$ using Refined UCT (Eq. \ref{eq:final_ucb})
    \STATE $\{\tau_m\}_{m=1}^M \leftarrow \text{EXPANDANDSIMULATE}(\pi_{\theta}, L)$ 

    \FOR{each trajectory $\tau_m$}
        \STATE $R_m \leftarrow \text{SIMULATETOGOAL}(\tau_m)$
        \STATE \text{BACKPROPAGATE}($\tau_m, R_m$) \COMMENT{Update $Q$-values}
    \ENDFOR

    \STATE \textit{// Fine-Grained Advantage Estimation}
    \STATE $A_{\text{all}} \leftarrow [ \ ]$
    \FOR{each $\tau_m = \{s_{m,1}, \dots, s_{m,T}\}$}
        \FOR{$n = 1$ \TO $T$}
            \STATE $\mathcal{S}_{\text{sib}} \leftarrow \text{GETSIBLINGS}(s_{m,n}) $
            \STATE $\bar{Q}_{\text{sib}} \leftarrow \text{MEAN} \{ Q(s') \mid s' \in \mathcal{S}_{\text{sib}} \}$
            \STATE $A_{m,n} \leftarrow Q(s_{m,n}) - \bar{Q}_{\text{sib}}$ \COMMENT{Node-level adv.}
        \ENDFOR
        \STATE $A(\tau_m) \leftarrow \frac{1}{T} \sum_{n=1}^{T} A_{m,n}$ \COMMENT{Trajectory adv.}
        \STATE $A_{\text{all}}.\text{APPEND}(A(\tau_m))$
    \ENDFOR
    
    \STATE $\pi_{\theta} \leftarrow \text{UPDATEPOLICY}(\pi_{\theta}, A_{\text{all}})$ \COMMENT{GRPO Loss Update}
\ENDFOR
\RETURN $\pi_{\theta}$
\end{algorithmic}
\end{algorithm}

\paragraph{Fine-Grained Advantage Function for Precise Policy Updates} In standard policy optimization frameworks like GRPO, the advantage function is typically computed based on the cumulative return of an entire trajectory. This coarse-grained signal poses a significant credit assignment challenge, as it fails to disambiguate the individual contributions of intermediate states. Consequently, it is difficult for the model to pinpoint which specific choices are most critical for achieving success.

To overcome this limitation, we introduce a fine-grained advantage function calculated at the intermediate state level. Our approach is tailored for a tree-search process wherein a LLM generates a set of $M$ candidate sequences (or completions), $\{\tau_1, \dots, \tau_M\}$, for a given planning problem. Each trajectory $\tau_m$ is composed of a sequence of intermediate states, $\tau_m = (s_{m,1}, s_{m,2}, \dots, s_{m,T_m})$.

Let $s_{m,n}$ be the $n$-th intermediate state in the $m$-th generated sequence. We estimate its corresponding state-value, or Q-value $Q_{m,n}$, as the mean empirical return from all Monte Carlo simulations that traverse this state. Specifically, if $W_{m,n}$ is the sum of cumulative rewards from all visits to state $s_{m,n}$ and $N_{m,n}$ is its total visit count, the Q-value is given by:
\begin{equation}
    Q_{m,n} = \frac{W_{m,n}}{N_{m,n}}
    \label{eq:q_value}
\end{equation}

We then define the advantage of a specific state, $A_{m,n}$, relative to its "sibling" states---i.e., the set of other candidate states $\{s_{j,n}\}_{j=1}^M$ that share a common prefix sequence. The state-level advantage is formulated as:
\begin{equation}
    A_{m,n} = Q_{m,n} - \mathrm{Mean}(Q_{\mathrm{siblings}})
    \label{eq:state_advantage}
\end{equation}
where the second term represents the mean Q-value across all sibling states at depth $n$. This formulation directly quantifies how much better the choice leading to $s_{m,n}$ is compared to the average of alternative choices at that decision point. A deliberate design choice is the omission of reward normalization (i.e., skipping division by the standard deviation). As the LLM often generates identical optimal completions, forgoing normalization prevents "reward hacking," where the value of a superior path could be artificially deflated due to its high frequency of generation. The necessity of this non-normalization approach in such scenarios has been formally demonstrated in \citep{liu2025understandingr1zeroliketrainingcritical}.

Finally, to align with the GRPO framework, we compute a single advantage value for each trajectory by averaging the advantages of all its constituent intermediate states. For a trajectory $\tau_m$ of length $T_m$, its overall advantage $A(\tau_m)$ is calculated as:
\begin{equation}
    A(\tau_m) = \frac{1}{T_m} \sum_{n=1}^{T_m} A_{m,n}
    \label{eq:traj_advantage}
\end{equation}
This trajectory-level advantage $A(\tau_m)$ is then used as the training signal within the GRPO loss function. This fine-grained approach to advantage calculation provides a more precise and informative signal, enabling the model to learn not only which overall sequences are effective, but also to discern the value of the specific intermediate states that are most critical for constructing an optimal plan. We present the full pseudo-code in Algorithm \ref{alg:m-grpo-functional}

\section{Experiments}
\subsection{Experimental Setup}
Given that spatial reasoning in embodied intelligence often operates under stringent computational constraints, we focus on open-source models with manageable parameter scales. Specifically, we employ Qwen3-4B-Instruct-2507 as our base model, which is further optimized via our M-GRPO framework to serve as the high-level planner. The untrained version of the same model served as both the environment decomposition model and the action generation model. We deliberately exclude large-scale proprietary closed-source models as they are often unsuitable for real-time embodied tasks, where low-latency local execution, robust offline functionality, and strict data privacy are paramount. Furthermore, to ensure our framework's generalizability across different open-source architectures, we evaluate it on the DeepSeek-R1-Distill-Llama-8B model family; due to space limits, these detailed results are provided in Appendix~\ref{sec:appendix_deepseek}.

\paragraph{Datasets} We evaluate our HSRL framework across four planning benchmarks with increasing difficulty to assess its performance and generalization capability. First, we use Maze Navigation\citep{NEURIPS2023_efb2072a}, a classical task consisting of 1,090 $10 \times 10$ grids. To create highly challenging scenarios, we configured 40\% of the cells in each map as obstacles. The dataset is partitioned into 668 training and 422 testing instances. To validate the real-world applicability of our approach, inspired by \citep{zhao2025llmnavi}, we leverage the R2V dataset \citep{liu2017raster}, which comprises 815 authentic architectural floorplans. We randomly selected a subset of 50 maps from this collection. Each floorplan was converted into a textual representation and scaled proportionally to a $20 \times 20$ resolution, followed by the random sampling of three pairs of start and goal positions for each layout. This procedure resulted in the evaluation benchmark for real-world indoor navigation tasks. Second, to assess out-of-distribution (OOD) generalization, we employ the Blocksworld benchmark\citep{system1-x}, whose test set is intentionally more complex, featuring more blocks and requiring longer plans (7--10 steps) than the training set (1--6 steps). Finally, we validate our framework on the novel and highly challenging GameTraversalBenchmark (GTB)\citep{gametravelbenchmark}. This benchmark contains 150 diverse maps with multiple objectives and paths exceeding 100 steps. As GTB lacks a training set, we evaluate our Maze-trained model in a zero-shot transfer setting to test its capabilities on complex, unseen tasks.


\paragraph{Baselines} We compare HSRL against a diverse set of representative baselines. First, we compare it with foundational reasoning strategies, including the classic Chain-of-Thought (CoT)\citep{NEURIPS2022_9d560961} and ReAct\citep{yao2023react}, which interleaves reasoning traces with actions for improved synergy. We also include advanced reasoning and self-reflection methods like Inner Monologue\citep{huang2023inner}, which enhances internal thought processes, and Reflexion\citep{shinn2023reflexion}, which uses iterative self-correction to refine plans. For direct planning, we use ProgPrompt\citep{singh2023progprompt} as a strong representative of in-context learning-based approaches. Furthermore, we contrast HSRL with search-based methods like Tree Planner\citep{hu2024treeplanner} and the hierarchical planner HyperTree\citep{guihypertree}, the latter of which is known to have limitations on spatial reasoning tasks. Finally, we include System-1.x\citep{system1-x}, a powerful baseline meticulously fine-tuned on tasks similar to ours, which employs a controller to switch between "fast-thinking" and "slow-thinking" modes.

\paragraph{Evaluation metrics} We evaluate our model's planning ability using metrics tailored to each benchmark. For the classical Maze and Blocksworld tasks, we measure the Completion Rate (CR), which is the percentage of successfully solved instances, and the Optimal Rate (OR), defined as the percentage of completed tasks where the plan length matches the shortest path computed by an A* search. For the more complex GameTraversalBenchmark (GTB), we adopt its official metrics. The primary metric is the GTB Score, a composite measure that assesses performance based on goal proximity, path length, and generation errors (see Appendix \ref{GTB:Details}). Additionally, we report Top-5 Accuracy, the fraction of tasks where the agent ends within five tiles of the target, to evaluate success in large-scale maps.

\subsection{Implementation Detail}
\paragraph{M-GRPO Finetuning.}
We fine-tuned the model using the trl library on eight NVIDIA A800 GPUs. 
The optimization was conducted using the AdamW optimizer with $\beta_1=0.9$ and $\beta_2=0.999$. 
We employed a learning rate of $1 \times 10^{-6}$ following a cosine decay schedule. 
The hyperparameters $\tau$ is 1.0 and $\gamma$ is $0.4$.
The Epoch Number was 1 and the Batch Size was 1. 
For the inference phase, the Temperature was set to 1.0 and the Num generations was 8. 


\begin{table*}[t]
\centering
\small
\caption{Main results: Comparison of HSRL against baseline methods across various spatial reasoning benchmarks. \textbf{CR} and \textbf{OR} denote Completion Rate and Optimality Rate, respectively. The best performance in each category is highlighted in \textbf{bold}.}
\label{tab:main_results}
\begin{tabular}{@{}lcccccccc@{}}
\toprule
\multirow{2}{*}{\textbf{Method}} & \multicolumn{2}{c}{\textbf{Maze (10$\times$10)}} & \multicolumn{2}{c}{\textbf{R2V (Real-World)}} & \multicolumn{2}{c}{\textbf{Blocksworld (5-7)}} & \multicolumn{2}{c}{\textbf{GTB}} \\ \cmidrule(lr){2-3} \cmidrule(lr){4-5} \cmidrule(lr){6-7} \cmidrule(lr){8-9}
 & CR (\%) $\uparrow$ & OR (\%) $\uparrow$ & CR (\%) $\uparrow$ & OR (\%) $\uparrow$ & CR (\%) $\uparrow$ & OR (\%) $\uparrow$ & Score $\uparrow$ & Top-5 Acc. $\uparrow$ \\ \midrule
Direct Answer & 23.69 & 23.45 & 1.33 & 1.13 & 6.50 & 6.00 & 23.61 & 25.96 \\
CoT & 43.12 & 38.39 & 16.67 & 12.33 & 10.50 & 8.00 & 26.58 & 31.61 \\
Reflexion & 45.02 & 37.91 & 22.00 & 21.33 & 15.00 & 8.50 & 29.34 & 37.76 \\
ReAct & 53.80 & 26.06 & 16.00 & 13.33 & 8.00 & 3.00 & 20.40 & 39.85 \\
ProgPrompt & 34.60 & 33.41 & 43.33 & 42.00 & 9.50 & 6.00 & 22.45 & 26.12 \\
Inner Monologue & 54.03 & 34.60 & 27.33 & 16.67 & 4.00 & 0.00 & 19.18 & 21.41 \\
System-1.x & 54.74 & 36.02 & 50.67 & 49.33 & 27.00 & 14.50 & 27.73 & 30.01 \\
HyperTree & 37.91 & 22.98 & 25.33 & 19.67 & 8.00 & 3.50 & 25.81 & 26.67 \\
Tree Planner & 39.10 & 27.01 & 6.33 & 3.00 & 7.00 & 4.00 & 25.28 & 25.46 \\ \midrule \addlinespace[0.5ex]
\textbf{HSRL (Ours)} & \textbf{61.37} & \textbf{47.87} & \textbf{62.67} & \textbf{55.33} & \textbf{30.50} & \textbf{21.00} & \textbf{32.69} & \textbf{44.41} \\ \bottomrule
\end{tabular}
\end{table*}

\paragraph{Reward Function.} To ensure training stability and prevent reward hacking, we implement a comprehensive reward function that weights task success against path length through multiple shaping terms. Specifically, to mitigate the tendency of RL agents to over-generate—a common phenomenon where models produce excessively long responses to maximize cumulative rewards—we introduce a stringent length penalty to encourage concise and effective reasoning. Detailed formulations of these composite reward functions are provided in Appendix \ref{M-GRPO:Reward}.

\subsection{Results Analysis}
The experimental results in Table~\ref{tab:main_results} clearly demonstrate the effectiveness of our framework.

\paragraph{HSRL Significantly Improves Task Performance.}
HSRL achieves state-of-the-art (SOTA) performance across all tasks. In the Maze ($10 \times 10$) task, it reaches a 61.37\% completion rate, markedly outperforming System-1.x (54.74\%) and ReAct (53.80\%). On the large-scale real-world R2V dataset, HSRL achieves a 62.67\% completion rate, maintaining high efficacy by decomposing long-horizon navigation to overcome the reasoning deterioration typical of monolithic methods. This advantage extends to the complex Blocksworld (30.50\% completion rate) and the GameTraversalBenchmark (GTB), where HSRL achieves the highest GTB Score (32.69 vs. Reflexion's 29.34) and Top-5 Accuracy (44.41\%).

\paragraph{Superior Solution Optimality.}
Generating optimal paths is a crucial metric of planning intelligence. HSRL achieves the highest optimality rates in Maze (47.87\%) and Blocksworld (55.33\%). By combining MCTS's forward-search with the LLM's general knowledge, our framework thoroughly explores the solution space, effectively avoiding local optima to devise concise and efficient paths.


\begin{table*}[t]
\centering
\small
\caption{Ablation study on the core components of our HSRL framework. We evaluate the contribution of each module across various benchmarks. \textbf{CR} and \textbf{OR} represent Completion Rate and Optimality Rate, respectively. The best performance is highlighted in \textbf{bold}.}
\label{tab:ablation}
\begin{tabular}{@{}lcccccccc@{}}
\toprule
\multirow{2}{*}{\textbf{Model Configuration}} & \multicolumn{2}{c}{\textbf{Maze (10$\times$10)}} & \multicolumn{2}{c}{\textbf{R2V (Real-World)}} & \multicolumn{2}{c}{\textbf{Blocksworld (5-7)}} & \multicolumn{2}{c}{\textbf{GTB}} \\ \cmidrule(lr){2-3} \cmidrule(lr){4-5} \cmidrule(lr){6-7} \cmidrule(lr){8-9}
 & CR (\%) $\uparrow$ & OR (\%) $\uparrow$ & CR (\%) $\uparrow$ & OR (\%) $\uparrow$ & CR (\%) $\uparrow$ & OR (\%) $\uparrow$ & Score $\uparrow$ & Top-5 Acc. $\uparrow$ \\ \midrule
State-Hierarchical Only & 50.24 & 13.03 & 52.33 & 25.67 & 10.00 & 9.50 & 26.16 & 29.85 \\
HSRL (Untrained) & 54.50 & 14.22 & 60.33 & 28.00 & 12.50 & 9.50 & 26.96 & 32.38 \\
HSRL (w/o MCTS) & 55.21 & 45.97 & 58.00 & 54.67 & 28.00 & 15.00 & 27.80 & 38.09 \\
HSRL (Standard UCT) & 60.43 & 46.44 & 59.33 & 55.00 & 29.50 & 18.00 & 30.65 & 40.29 \\ \midrule \addlinespace[0.5ex]
\textbf{HSRL (Ours)} & \textbf{61.37} & \textbf{47.87} & \textbf{62.67} & \textbf{55.33} & \textbf{30.50} & \textbf{21.00} & \textbf{32.69} & \textbf{44.41} \\ \bottomrule
\end{tabular}
\end{table*}

\paragraph{Cross-Task Robustness and Generalization.}
The value of a general-purpose planning model lies in its cross-task generalization capability. As shown in the table, our method performs exceptionally well on the classical spatial reasoning tasks of Maze and Blocksworld, and the complex world knowledge required for the Game Travel Benchmark. In contrast, some baselines exhibit strong task-specific biases; for example, while ReAct performs reasonably well in Maze, its completion rate plummets to just $3.00\%$ in Blocksworld. This comparison validates the robustness of our hierarchical framework, which consistently decomposes complex problems into manageable sub-goals for effective problem-solving, irrespective of the task modality. Moreover, our excellent performance in Blocksworld demonstrates strong out-of-distribution generalization capabilities.












\subsection{Further Analysis}

\paragraph{Ablation Study.}
To validate the contribution of each component, we conducted a comprehensive ablation study (Table~\ref{tab:ablation}). The results reveal a clear hierarchy of importance. Replacing the modified UCT with the standard UCT significantly degrades performance on the challenging GTB task. Removing the MCTS module—thereby reverting M-GRPO to a standard GRPO training paradigm (HSRL w/o MCTS)—leads to a notable decline in both success and optimality, confirming that its systematic, forward-looking search is crucial for exploring diverse solution pathways. Further removing the M-GRPO policy optimization, where the base model is used directly as the high-level planner without further training (HSRL Untrained) causes a precipitous performance collapse, especially in optimality (e.g., Maze optimality plummets from 47.87\% to 45.97\%). This demonstrates that M-GRPO is the core engine that translates the rich search experience from MCTS into a refined planning intuition, endowing the LLM with the ability to generate high-quality, task-aligned sub-goals. Finally, the performance of the State-Hierarchical Only configuration—which eliminates environment-level hierarchy and relies on the base model for both intermediate waypoint identification and low-level action generation—still significantly surpasses direct answering methods, and the inclusion of environment-hierarchical approaches effectively improves task completion.

\paragraph{Fair Comparison.}
To ensure a fair evaluation and address concerns regarding unequal computational budgets, we validate our framework across both inference-only and training-based settings. In the inference-only track, our zero-shot "HSRL (Untrained)" relies solely on the proposed state and environment decomposition prompting, yet it achieves a 54.50\% completion rate on the Maze task, significantly outperforming strong baselines like CoT (43.12\%), ProgPrompt (34.60\%), and ReAct (53.80\%). Furthermore, in the training-based track, we compare HSRL against System-1.x under identical training datasets and interaction budgets. Under these strictly controlled conditions, HSRL still yields substantial improvements. These dual-track results confirm that our performance gains stem fundamentally from the architectural advantages of hierarchical decomposition and the superior optimization logic of M-GRPO, rather than computational disparities.

\begin{figure}[h!]
    
    \begin{minipage}{0.48\textwidth}
        \centering
        \includegraphics[width=\textwidth]{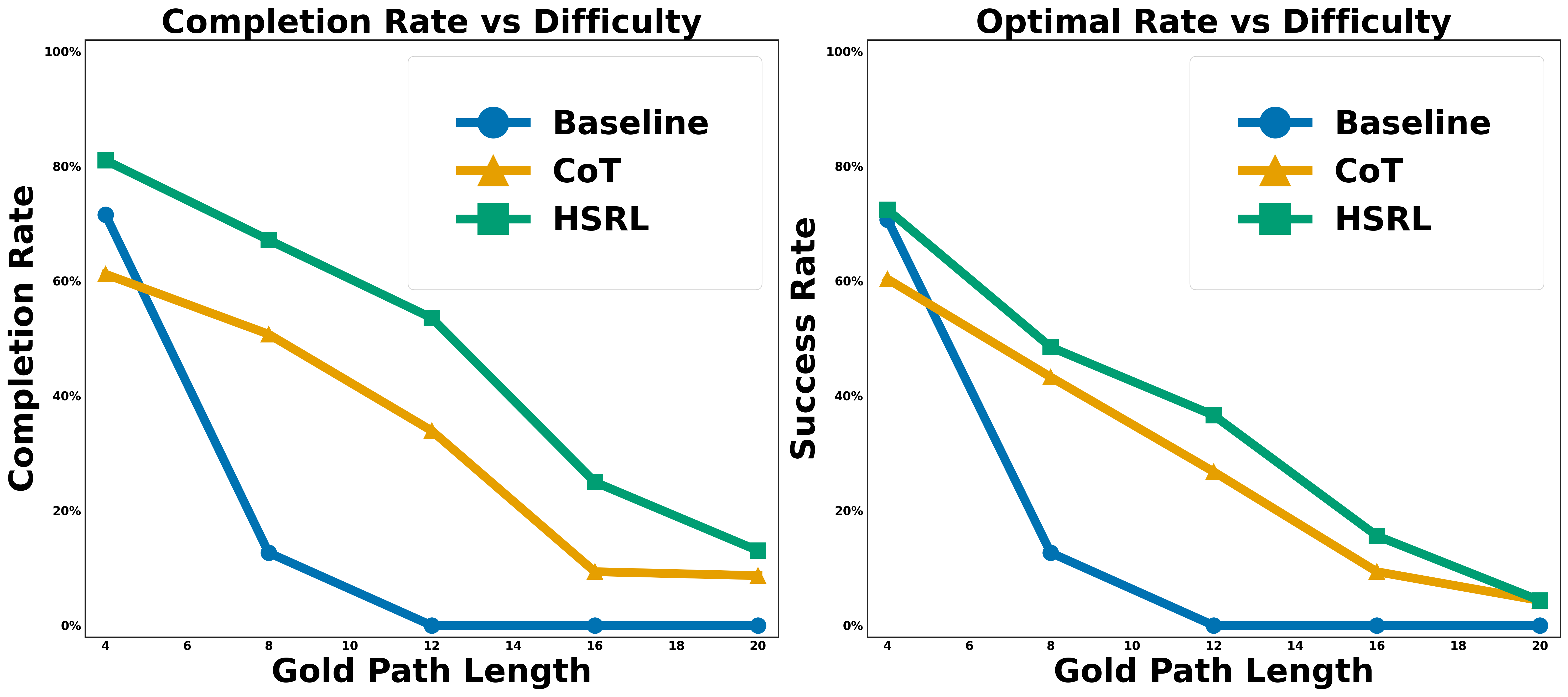}
        \captionof{figure}{Performance degradation as task difficulty increases. HSRL shows greater robustness.}
        \label{fig:difficulty}
    \end{minipage}
    \hfill 
    \begin{minipage}{0.48\textwidth}
        \centering
        \includegraphics[width=\textwidth]{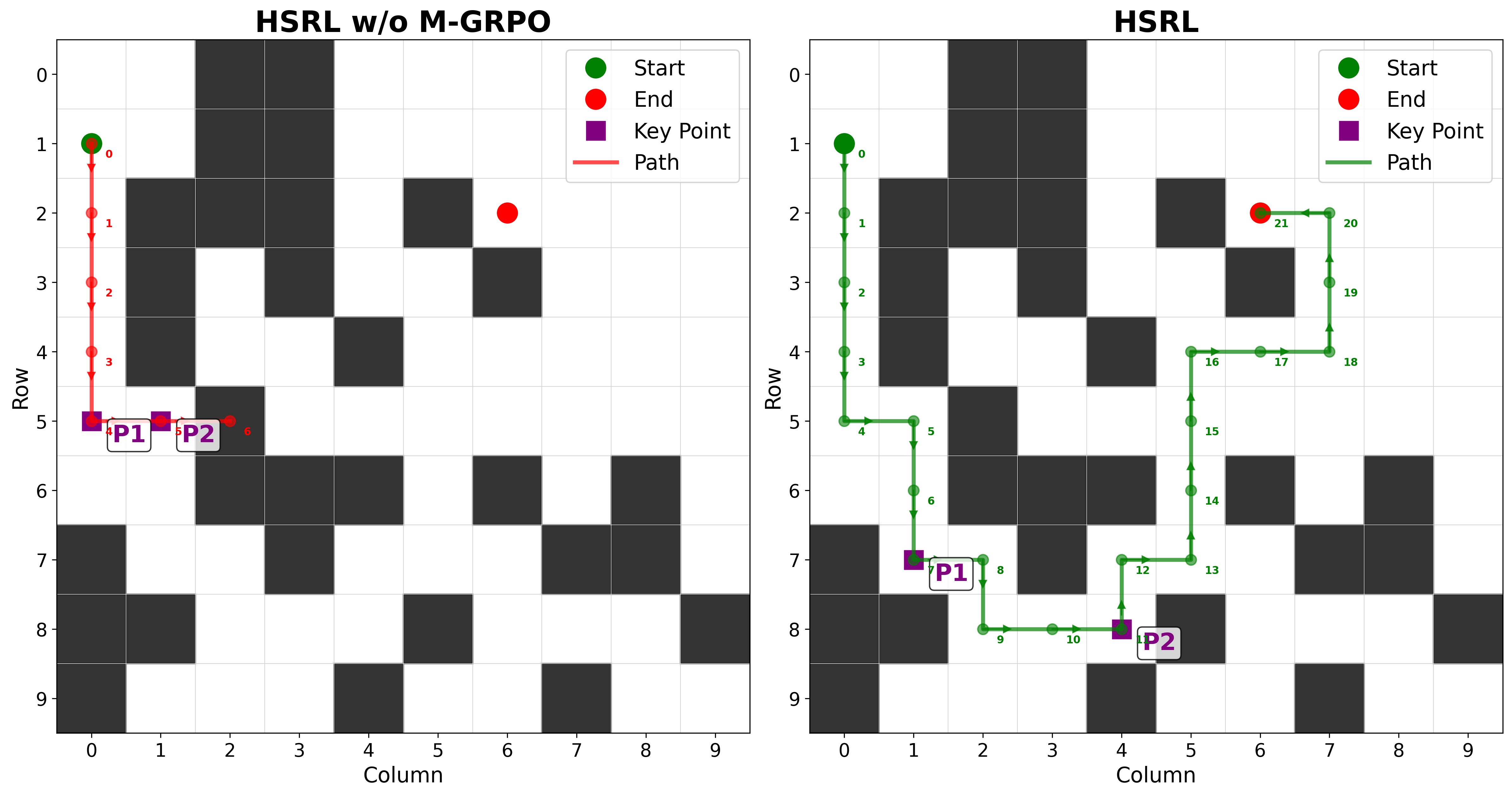}
        \captionof{figure}{Qualitative comparison of a generated plan before (left) and after (right) M-GRPO training.}
        \label{fig:case_study}
    \end{minipage}

\end{figure}

\paragraph{Robustness and Qualitative Insights.}
As shown in Figure~\ref{fig:difficulty}, HSRL's performance degrades far more gracefully with increasing task difficulty than baselines, maintaining a substantial advantage on challenging Maze instances. This resilience stems fundamentally from improved high-level planning. Qualitatively (Figure~\ref{fig:case_study}), while an untrained model generates ill-conceived sub-goals causing failure, the M-GRPO trained HSRL produces a strategic path. This synergy of quantitative robustness and qualitative intelligence validates HSRL's effectiveness in complex, long-horizon planning.







\section{Conclusion}

This paper introduced HSRL, a hierarchical framework that overcomes LLM spatial reasoning limitations through state and environment decomposition. By combining MCTS-guided GRPO algorithm, HSRL achieves SOTA performance across navigation and strategic benchmarks. Our results show substantial gains in completion rates and path optimality over existing methods. This work establishes a scalable approach for integrating LLMs into complex physical environments and embodied intelligence tasks.

\section*{Limitations}
Despite the advancements in spatial reasoning achieved by this study, several limitations remain. First, M-GRPO introduces additional computational overhead during the training phase. Because integrating MCTS for policy optimization requires extensive path simulations, the training process is inherently more time-consuming than conventional fine-tuning. However, this computational cost is strictly confined to the offline training stage and does not compromise inference efficiency. Second, due to the scarcity of large-scale datasets featuring purely textual descriptions of 3D environments, our current evaluation focuses primarily on 2D indoor navigation tasks (e.g., R2V). Extending this framework to more complex 3D spatial representations and multimodal environments remains an important direction for future work.

Third, while HSRL demonstrates significant superiority in complex spatial reasoning, its applicability is intrinsically bounded by the nature of the task domain. Our framework relies heavily on spatial topology to identify geometric intermediate states, constructing localized sub-environments to mitigate the noise of tightly coupled actions. While this design is highly effective in continuous domains lacking distinct semantic boundaries, it implies that HSRL is not a universal planner. In tasks driven by abstract logic or general semantic planning—where sub-goals are loosely coupled and naturally divided by linguistic breaks—the geometric decomposition mechanism cannot be fully utilized. Future research could explore adaptive mechanisms that seamlessly switch between geometric spatial decomposition and traditional semantic decomposition based on specific task contexts.

\bibliography{custom}

\appendix
\section*{Appendix} 

\section{Related Work}
\label{gen_inst}
\subsection{Spatial Reasoning in large language models}
Many researchers have pointed out that LLMs have weaknesses in spatial reasoning or spatial planning\citep{aghzal2024can,aghzal2024look}. To address these issues, some methods leverage in-context examples and prompting techniques, such as Chain-of-Thought (CoT)\citep{NEURIPS2022_9d560961} and Tree-of-Thought (ToT)\citep{NEURIPS2023_271db992}, which have demonstrated remarkable reasoning abilities in various tasks. However, for spatial reasoning tasks, in-context learning often fails because LLMs lack spatial reasoning knowledge or their knowledge even conflicts with it.

To overcome this challenge, some studies utilize LLMs for general-purpose reasoning, converting spatial information into logical forms\citep{yang-etal-2023-coupling} or using them as a general pattern machine for sequence transformation\citep{pmlr-v229-mirchandani23a,gong2024evolutionarylargelanguagemodel}. Recently, other works have evaluated LLMs as a cognitive capability in navigation and planning tasks\citep{NEURIPS2023_dc9d5dcf}. However, these methods perform poorly in tasks requiring continuous action reasoning.

Another mainstream approach introduces closed-loop feedback mechanisms. Some works, like \citep{DBLP:journals/corr/abs-2405-06682}, use self-reflection for self-evaluation and replanning, while others adopt external feedback for reflection \citep{10.1145/3657604.3662042}. Furthermore, the Vision-of-Thought (VoT) method \citep{NEURIPS2024_a45296e8} materializes intermediate states to assist with reasoning. Nevertheless, this iterative feedback loop often results in high costs and inefficiency in querying or interactions.

\subsection{Hierarchical Method}
Hierarchical reasoning breaks down decision-making tasks into multiple levels, from high-level strategic planning to low-level specific control. This decomposition reduces computational complexity by solving several less difficult sub-tasks, thus enabling the handling of tasks more challenging than direct complex reasoning. Hierarchical reasoning has achieved notable results in many reinforcement learning tasks, especially in embodied AI scenarios. For example, \citep{10.1049/iet-its.2019.0317} has applied hierarchical methods to autonomous driving, allowing for smooth and safe decision-making on highways. \citep{lu2023action} and \citep{9834298} separate decision-making tasks into different layers, such as global path planning and local motion control. These models benefit from breaking down the decision-making process into simpler, more tractable components, enabling each layer to focus on a specific task. This enhances computational efficiency and decision accuracy in complex environments.

In recent years, hierarchical reasoning methods have also been successfully introduced into the planning tasks of LLMs. For instance, DeAR\citep{NEURIPS2024_01025a4e} imitates the human reasoning cycle by using a tree-based question decomposition approach to organize the reasoning process and break down problems into simpler sub-questions. HyperTree Planning\citep{gui2025hypertree} is a new paradigm that enhances LLM reasoning with a hypertree structure. It effectively breaks down intricate reasoning steps using a flexible divide-and-conquer strategy to handle diverse constraints and manage multiple distinct sub-tasks, demonstrating superior performance in complex tasks like travel planning. Plan-and-Act\citep{erdogan2025planandact} explicitly separates high-level planning from low-level execution. This framework includes a PLANNER model for generating structured high-level plans and an EXECUTOR model for translating these plans into environment-specific actions, thereby improving performance on complex multi-step tasks such as web navigation. 

However, these methods primarily focus on high-level, coarse-grained task planning, failing to fully leverage hierarchical reasoning for fine-grained, low-level spatial and motion control. Furthermore, while prior MCTS-based semi-online methods (e.g., SEEA-R1) have explored search-based planning, they are largely restricted to discrete action spaces and suffer from distribution shifts on static trees. To bridge these gaps, our study introduces M-GRPO to solve complex action planning problems. By employing strictly on-policy online optimization with dynamic iterative expansion, our method efficiently navigates the combinatorially explosive, high-dimensional search spaces inherent in spatial planning.

\begin{table*}[htbp]
\centering
\caption{Generalization results on the DeepSeek model across different spatial reasoning benchmarks. Bold denotes the best performance.}
\label{tab:deepseek_generalization}
\resizebox{\textwidth}{!}{%
\begin{tabular}{@{}lcccccccc@{}}
\toprule
\multirow{2}{*}{\textbf{Method}} & \multicolumn{2}{c}{\textbf{Maze (10$\times$10)}} & \multicolumn{2}{c}{\textbf{R2V (Real-World)}} & \multicolumn{2}{c}{\textbf{Blocksworld (5-7)}} & \multicolumn{2}{c}{\textbf{GTB}} \\ \cmidrule(lr){2-3} \cmidrule(lr){4-5} \cmidrule(lr){6-7} \cmidrule(lr){8-9} 
 & CR (\%) $\uparrow$ & OR (\%) $\uparrow$ & CR (\%) $\uparrow$ & OR (\%) $\uparrow$ & CR (\%) $\uparrow$ & OR (\%) $\uparrow$ & Score $\uparrow$ & Top-5 Acc. $\uparrow$ \\ \midrule
Direct Answer & 11.85 & 8.77 & 2.67 & 2.67 & 8.50 & 8.50 & 20.76 & 22.41 \\
CoT & 13.50 & 13.03 & 10.67 & 8.67 & 13.50 & 12.50 & 24.11 & 24.59 \\
System-1.x & 18.72 & 15.40 & 20.67 & 19.33 & 29.50 & 23.00 & 25.89 & 32.97 \\ \midrule \addlinespace[0.5ex]
\textbf{HSRL (Ours)} & \textbf{23.70} & \textbf{20.38} & \textbf{28.00} & \textbf{22.67} & \textbf{33.00} & \textbf{25.00} & \textbf{27.92} & \textbf{34.82} \\ \bottomrule
\end{tabular}%
}
\end{table*}
\section{Generalization on DeepSeek Model}
\label{sec:appendix_deepseek}

To further evaluate the architecture-agnostic nature and generalization capability of our HSRL framework, we conduct additional experiments using the DeepSeek-V3 model as the backbone. We maintain the same experimental settings and evaluation metrics (CR, OR, Score, and Top-5 Acc.) across four diverse spatial reasoning benchmarks: Maze, R2V, Blocksworld, and GTB.

As shown in Table~\ref{tab:deepseek_generalization}, HSRL consistently outperforms other competitive baselines when integrated with the DeepSeek architecture. Notably, on the Blocksworld (5-7) task, HSRL achieves a CR of 33.00\% and an OR of 25.00\%, marking a significant improvement over the Direct Answer and CoT methods. These results demonstrate that the hierarchical task decomposition and M-GRPO optimization in our framework are not biased toward a specific model family (e.g., Qwen) but provide a robust enhancement for spatial planning across different Large Language Models.


\section{More Background}
\subsection{GTB Score}\label{GTB:Details}

\begin{equation}
\begin{aligned}
    \text{GTB\_Score} = \frac{1}{M} \sum_{m=1}^{M} \frac{1}{\Delta R^{(m)}} \Bigl( & R^{(m)} \\ - \text{LLM}_{PL}^{(m)} - \varepsilon^{(m)} - R_{\min}^{(m)} \Bigr)
\end{aligned}
\end{equation}

\noindent
where:
\begin{itemize}
    \item $M$ = total number of maps in the dataset.
    \item $R^{(m)}$ = reward obtained for map $m$, determined by the final distance $d$ to the objective:
    \[
    R^{(m)} =
    \begin{cases}
        +200, & d = 0 \\
        +100, & d = 1 \\
        +50,  & d \in [2,3] \\
        +25,  & d \in [3,5] \\
        -50,  & d \in [5,8] \\
        -100, & d \geq 8
    \end{cases}
    \]
    \item $\text{LLM}_{PL}^{(m)}$ = path length taken by the LLM agent on map $m$.
    \item $\varepsilon^{(m)}$ = total generation errors made by the LLM on map $m$.
    \item $R_{\max}^{(m)} = 200 - A^{*}_{PL}(m)$, the maximum achievable reward (perfect path with no errors), 
          where $A^{*}_{PL}(m)$ is the optimal path length computed by an A* agent.
    \item $R_{\min}^{(m)} = -100 - A^{*}_{PL}(m) - \varepsilon_{\max}^{(m)}$, the minimum achievable reward 
          (farthest position, maximal path cost, and maximal errors).
\end{itemize}

\section{MORE IMPLEMENTATION DETAILS}
\subsection{M-GRPO Reward}\label{M-GRPO:Reward}
For a sampled completion $completion_i$, we parse its anchor list 
$A_i = [a_{i1}, a_{i2}, \ldots, a_{in}]$. 
If the anchor list cannot be parsed, we directly assign a fixed penalty; 
otherwise, the reward score is computed by a signed power transformation to enlarge 
the margin between high- and low-quality completions:
\begin{equation}
\label{eq:reward_1}
R_i =
\begin{cases}
\texttt{PARSE\_FAIL\_PENALTY}, & \text{if } A_i = \emptyset, \\[6pt]
\mathrm{sign}(z_i)\,|z_i|^{p}, & \text{otherwise},
\end{cases}
\end{equation}
where $\mathrm{sign}(z_i)$ preserves the direction of $z_i$, and $p>1$ 
amplifies its magnitude non-linearly.

The raw score $z_i$ aggregates the quality of anchors visited by a trajectory, 
while discouraging the use of overly many anchors through a penalty term:
\begin{equation}
\label{eq:reward_2}
z_i = \sum_{a \in A_i} \overline{r}(a) 
- \alpha \cdot \max \!\bigl(0,\,|A_i| - A_{\text{expected}}\bigr).
\end{equation}

To evaluate each anchor consistently, we first assign each completion an initial reward $r_i$ according to its alignment with the Manhattan distance of the optimal $A^*$ path. :
\begin{equation}
\begin{split}
r_i = BASIC\_QUALITY\_SCORE - \\ \bigl | \sum\limits_{a_i\in A_i}|a_{i+1}-a_i|-\sum\limits_{\hat{a_i}\in A^*}|\hat{a_{i+1}}-\hat{a_i}| \bigl |.
\end{split}
\end{equation}

Each anchor reward $\overline{r}(a)$ is then defined as the average quality of 
all completions that pass through it, reflecting a consensus measure across 
different trajectories:
\begin{equation}
\overline{r}(a) = 
\frac{\sum\limits_{i:\, a \in A_i} r_i}{|\{\, i \mid a \in A_i \,\}|}.
\end{equation}

\section{Declaration of Use of AI Assistants}\label{prompt}
During the preparation of this work, the authors used Gemini-3-Flash to polish the manuscript for language clarity and to assist with LaTeX formatting. Following the use of this tool, the authors reviewed and edited the content as needed and take full responsibility for the final version of the paper.

\section{Quantitative Analysis of Computational Costs}
\label{app:computational_costs}

\paragraph{Training Time Comparison.} Table~\ref{tab:training_time} reports the training time for Standard GRPO and our M-GRPO under identical hardware and batch size configurations.

\begin{table}[h]
\centering
\caption{Training time comparison (same hardware and batch size).}
\label{tab:training_time}
\begin{tabular}{lcc}
\toprule
\textbf{Environment} & \textbf{Standard GRPO} & \textbf{M-GRPO (Ours)} \\
\midrule
Maze        & 1h 35min  & 4h 49min \\
Blocksworld & 29min     & 2h 05min \\
\bottomrule
\end{tabular}
\end{table}

\paragraph{Inference Efficiency.} Crucially, inference remains highly efficient and comparable to standard Chain-of-Thought (CoT). Table~\ref{tab:inference_time} presents the average inference time per task.

\begin{table}[H]  
\centering
\caption{Average inference time comparison.}
\label{tab:inference_time}
\begin{tabular}{lcc}
\toprule
\textbf{Task} & \textbf{CoT} & \textbf{HSRL} \\
\midrule
Blocksworld & 19.20s & 20.18s \\
Maze        & 22.34s & 28.23s \\
\bottomrule
\end{tabular}
\end{table}

\section{Prompts and Examples}\label{prompt}


\begin{figure*}[ht] 
    \centering
    \includegraphics[width=0.95\textwidth]{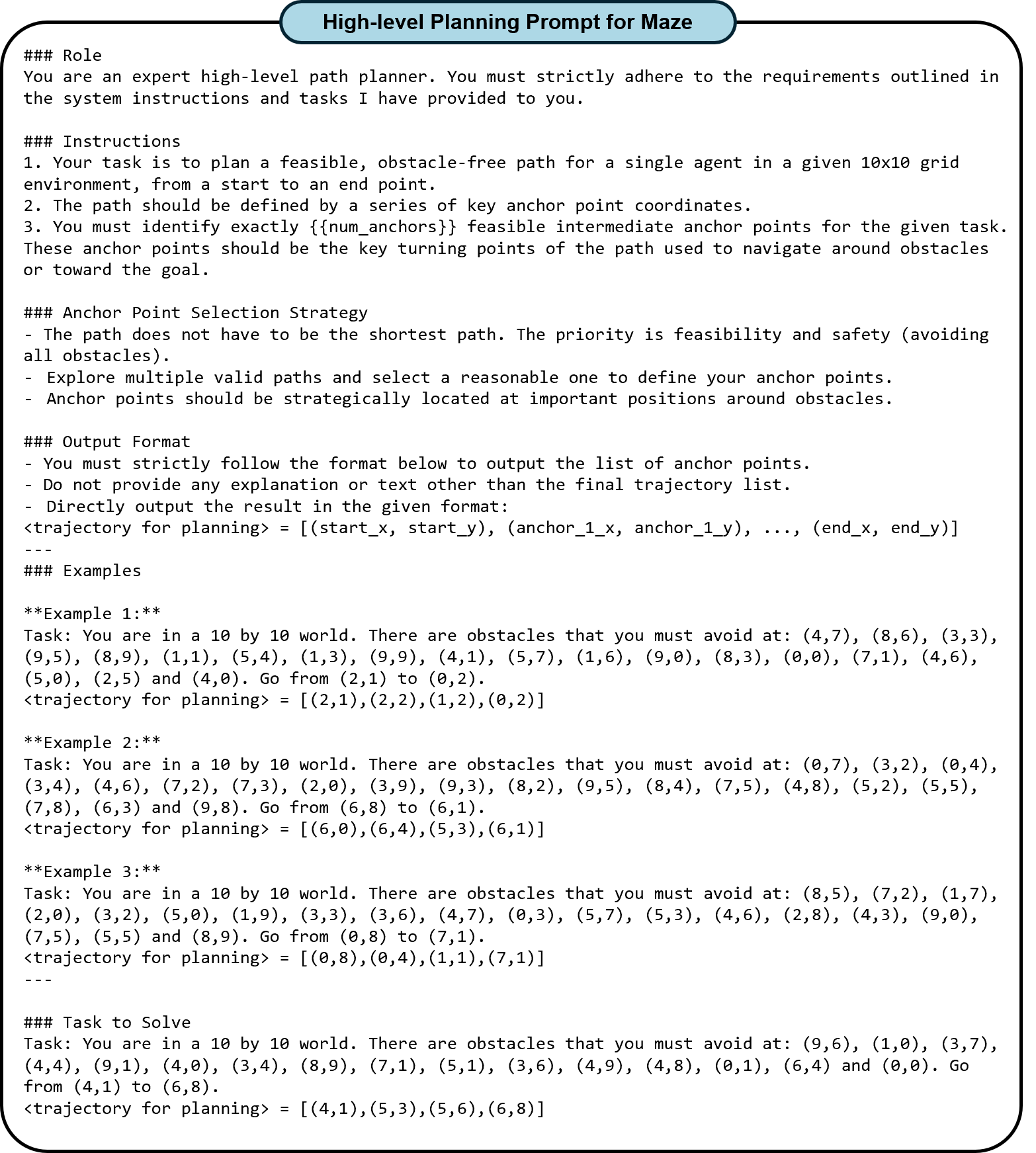}
    \label{fig:prompt_1}
    \vspace{1.5em}
\end{figure*}

\begin{figure*}[ht] 
    \centering
    \includegraphics[width=0.95\textwidth]{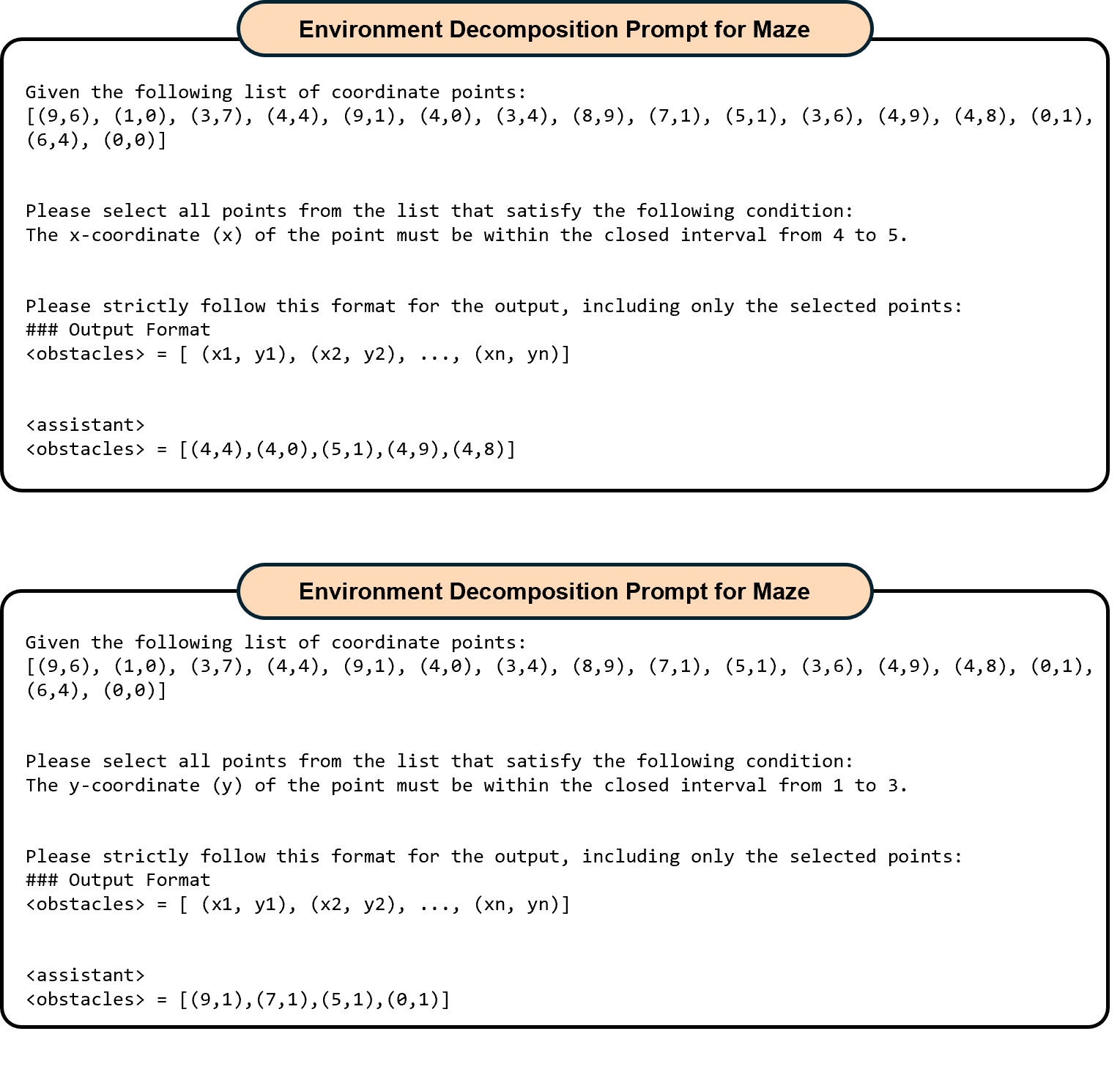}
    \label{fig:prompt_2}
    \vspace{1.5em}
\end{figure*}

\begin{figure*}[ht] 
    \centering
    \includegraphics[width=0.95\textwidth]{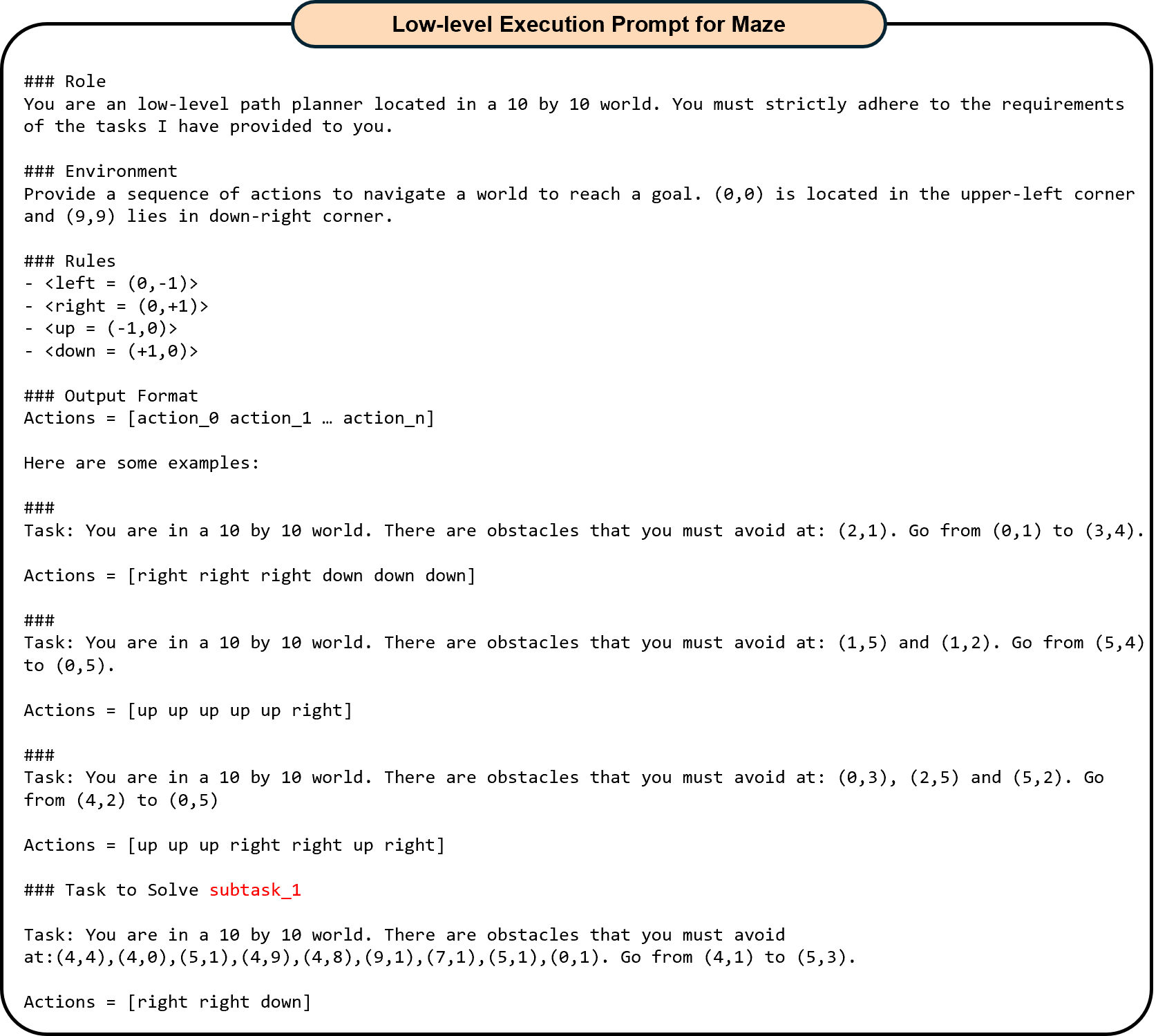}
    \label{fig:prompt_3}
    \vspace{1.5em}
\end{figure*}

\begin{figure*}[ht] 
    \centering
    \includegraphics[width=0.95\textwidth]{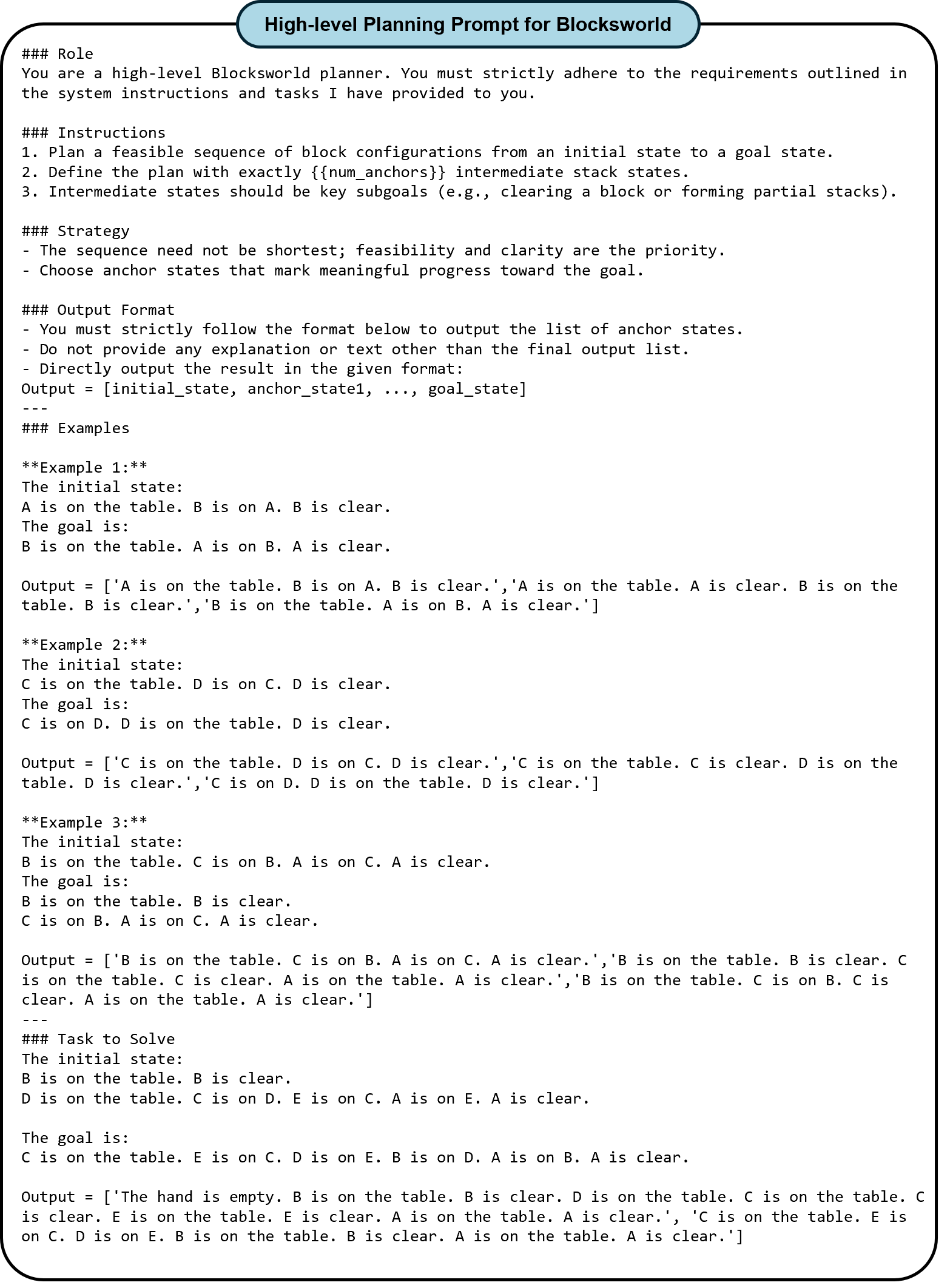}
    \label{fig:prompt_4}
    \vspace{1.5em}
\end{figure*}

\begin{figure*}[ht] 
    \centering
    \includegraphics[width=0.95\textwidth]{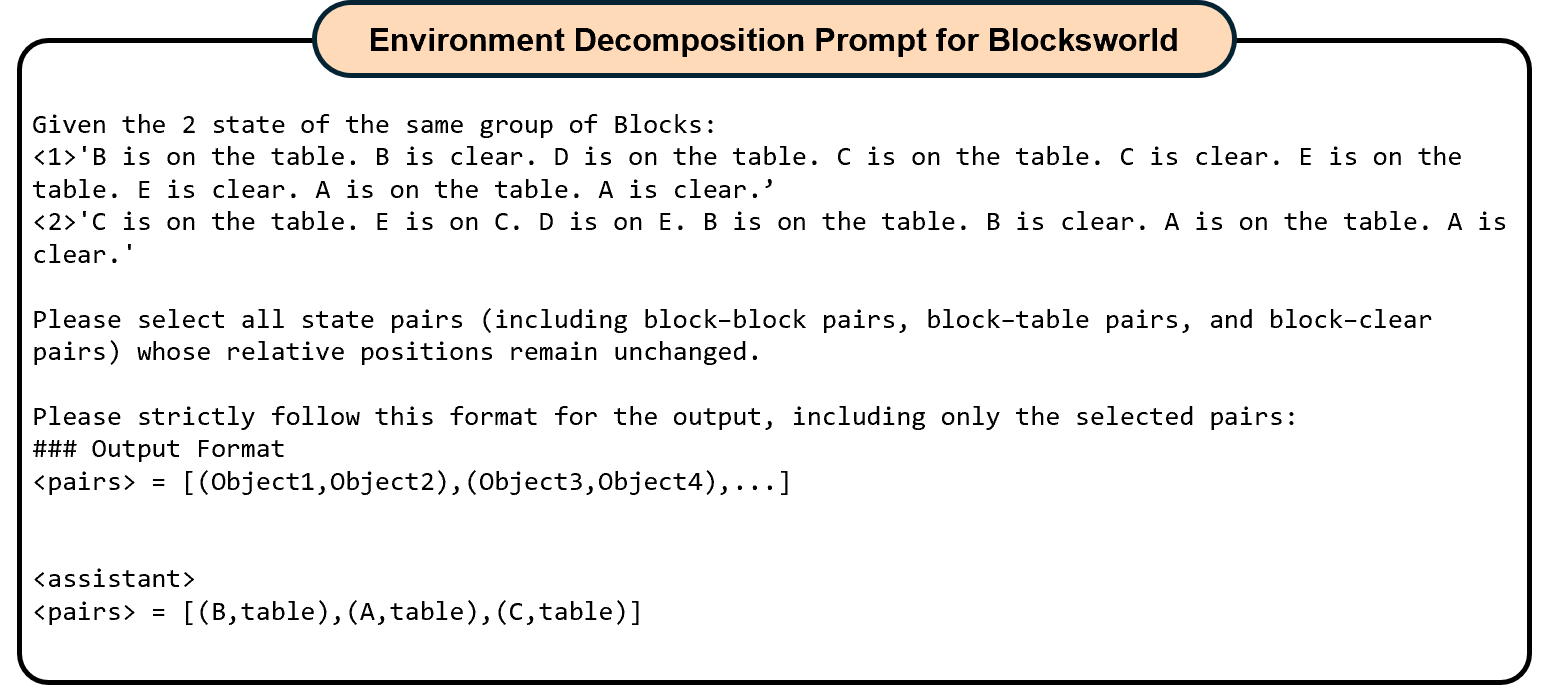}
    \label{fig:prompt_5}
    \vspace{1.5em}
\end{figure*}

\begin{figure*}[ht] 
    \centering
    \includegraphics[width=0.95\textwidth]{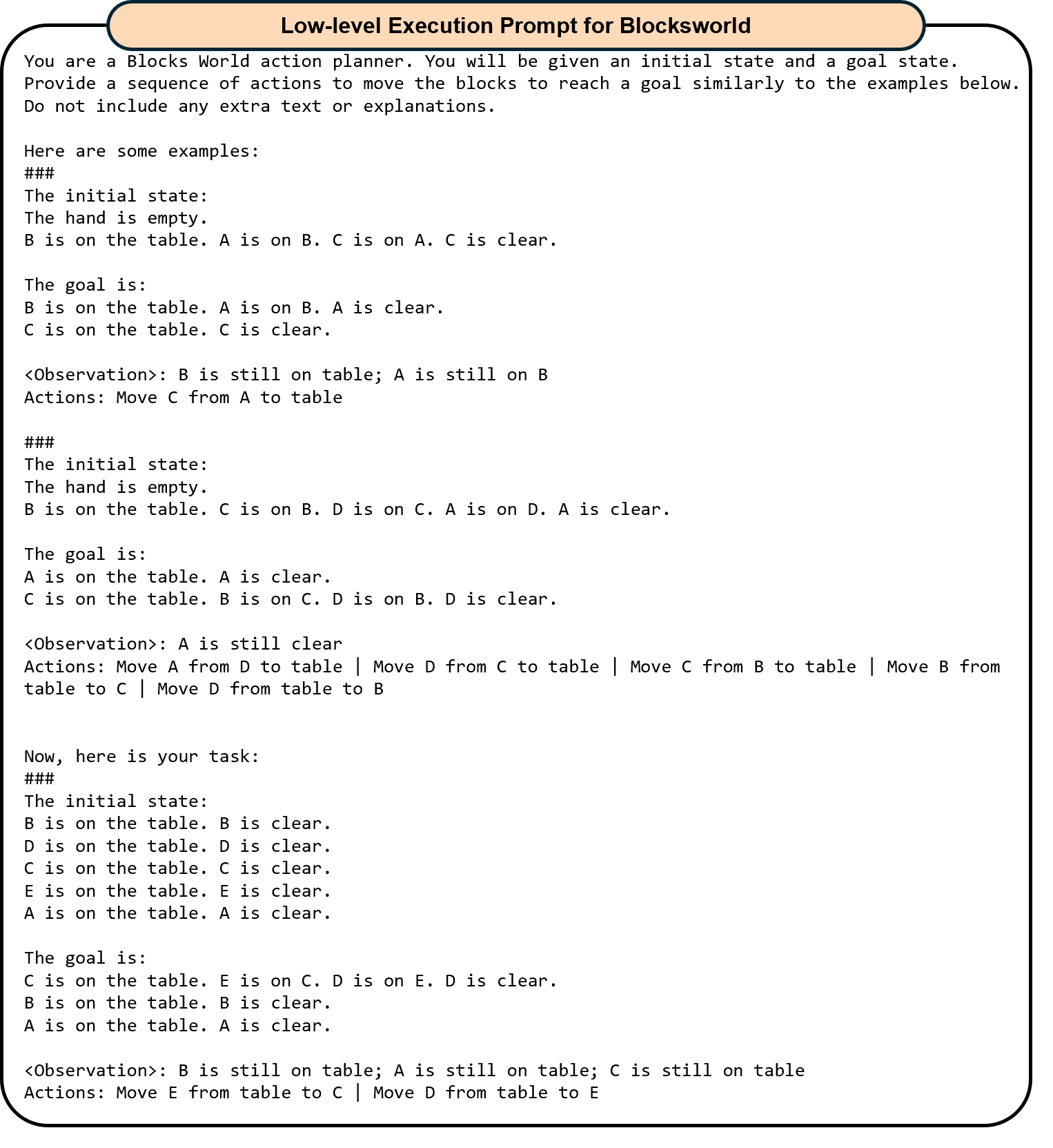}
    \label{fig:prompt_6}
    \vspace{1.5em}
\end{figure*}

\begin{figure*}[ht] 
    \centering
    \includegraphics[width=0.95\textwidth]{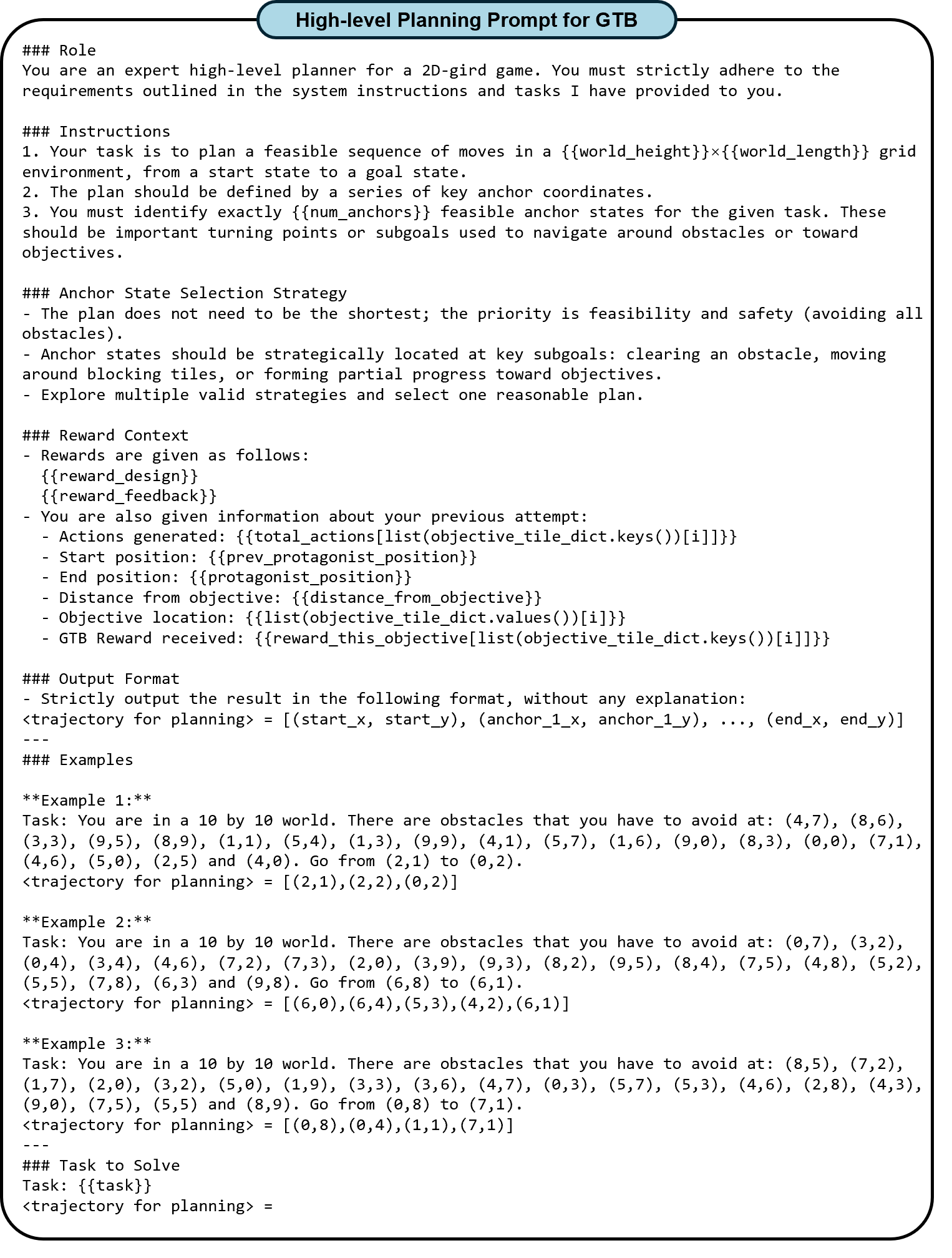}
    \label{fig:prompt_7}
    \vspace{1.5em}
\end{figure*}

\begin{figure*}[ht] 
    \centering
    \includegraphics[width=0.95\textwidth]{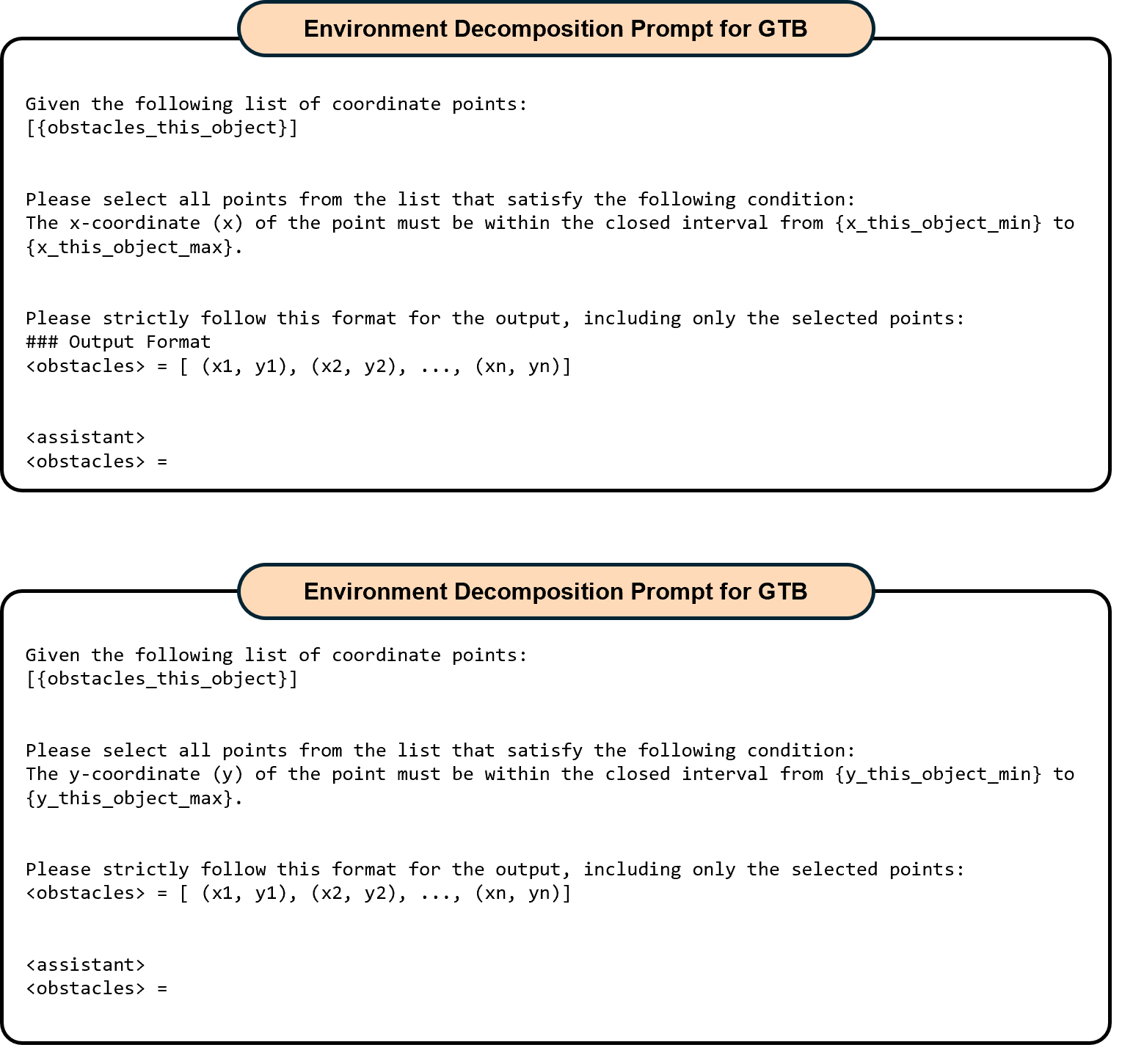}
    \label{fig:prompt_8}
    \vspace{1.5em}
\end{figure*}

\begin{figure*}[ht] 
    \centering
    \includegraphics[width=0.95\textwidth]{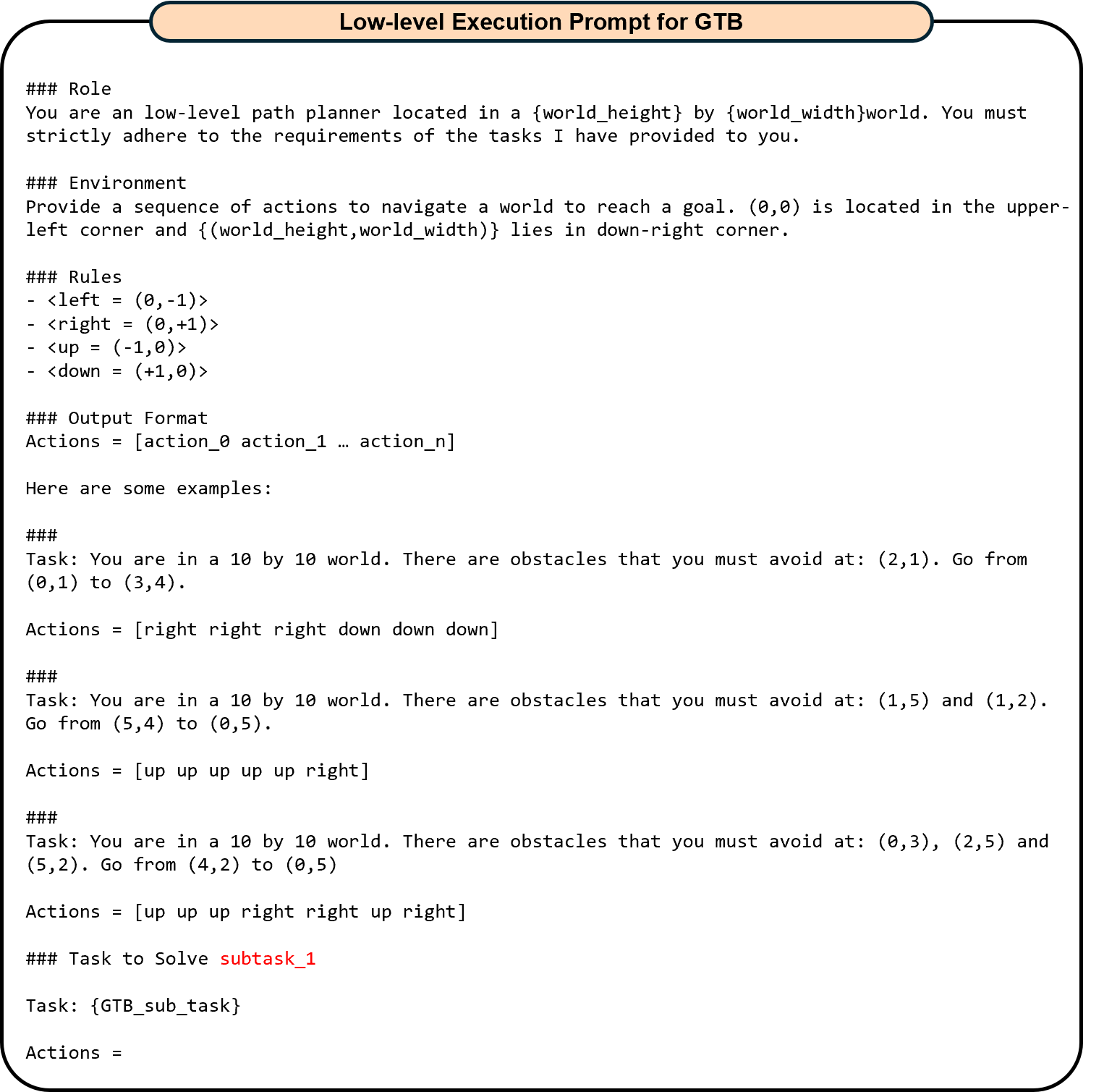}
    \label{fig:prompt_9}
    \vspace{1.5em}
\end{figure*}


\end{document}